%% file: root.tex
\documentclass[letterpaper, 10 pt, conference]{ieeeconf}  

\IEEEoverridecommandlockouts                              

\usepackage[noadjust]{cite}

\usepackage{graphicx} 
\usepackage{amsmath} 
\usepackage{amssymb}  
\usepackage{float}
\usepackage{hyperref}
\usepackage{color}
\usepackage{algorithm}
\usepackage{algpseudocode}
\usepackage{subcaption}
\usepackage{xfrac}

\usepackage{booktabs, multirow} 
\usepackage{soul}
\let\labelindent\relax
\usepackage{enumitem}

\usepackage[table]{xcolor} 
\usepackage{changepage,threeparttable} 
\definecolor{international_orange}{RGB}{240, 74, 0}

\newcommand{\units}[1]{\text{#1}}
 
\title{\LARGE \bf
DefGraspNets: Grasp Planning on 3D Fields with Graph Neural Nets 
}

\author{Isabella Huang$^{1}$, Yashraj Narang$^{2}$, Ruzena Bajcsy$^{1}$, Fabio Ramos$^{2,3}$, Tucker Hermans$^{2,4}$, Dieter Fox$^{2,5}$
\thanks{$^{1}$Department of Electrical Engineering and Computer Sciences, University of California, Berkeley, USA ;$^{2}$NVIDIA Corporation, Seattle, USA;$^{3}$School of Computer Science, University of Sydney, Sydney, Australia;$^{4}$School of Computing, University of Utah, Salt Lake City, USA;$^{5}$Paul G. Allen School of Computer Science \& Engineering, University of Washington, Seattle, USA}}
\usepackage[skip=1pt,font=small,labelfont=bf]{caption}
\begin{document}

\maketitle
\thispagestyle{empty} 
\pagestyle{empty}

\begin{abstract}
  Robotic grasping of 3D deformable objects is critical for real-world applications such as food handling and robotic surgery. Unlike rigid and articulated objects, 3D deformable objects have infinite degrees of freedom. Fully defining their state requires 3D  deformation and stress fields, which are exceptionally difficult to analytically compute or experimentally measure. Thus, evaluating grasp candidates for grasp planning typically requires accurate, but slow 3D finite element method (FEM) simulation. Sampling-based grasp planning is often impractical, as it requires evaluation of a large number of grasp candidates. Gradient-based grasp planning can be more efficient, but requires a differentiable model to synthesize optimal grasps from initial candidates.
  Differentiable FEM simulators may fill this role, but are typically no faster than standard FEM. In this work, we propose learning a predictive graph neural network (GNN), DefGraspNets, to act as our differentiable model.
  We train DefGraspNets to predict 3D stress and deformation fields based on FEM-based grasp simulations.
  DefGraspNets not only runs up to $1500$x faster than the FEM simulator, 
  but also enables fast gradient-based grasp optimization over 3D stress and deformation metrics. We design DefGraspNets to align with real-world grasp planning practices and demonstrate generalization across multiple test sets, including real-world experiments.
\end{abstract}


\input{1_introduction}
\input{2_related_work}
\input{3_model}
\input{4_training}
\input{5_grasp_planning}

\input{6_prediction_results}
\input{7_grasp_planning_results}
\input{8_ablation}

\input{10_discussion}

\clearpage
\newpage
\bibliographystyle{unsrt}
\bibliography{references}  

\end{document}

%% file: 1_introduction.tex
\section{Introduction}
Deformable objects are omnipresent in our world, and grasping them is critical for food handling \cite{Gemici2014IROS}, robotic surgery \cite{Smolen2009ICACHI}, and domestic tasks \cite{Sanchez2018IJRR, Zhu2022RAM}. However, their physical complexities pose challenges for key aspects of grasp planning, including modeling, simulation, learning, and optimization. Deformable objects have infinite degrees of freedom and require continuum mechanics models to accurately predict their responses to body forces (e.g., gravity) and surface tractions (e.g., contacts). For deformable solids, continuum models can predict two field quantities critical for robot grasping, \textit{stress} tensors and \textit{deformation} vectors defined at every point in the object~\cite{timoshenko2010}. In general, low-stress grasps are desirable to reduce material fatigue from repeated grasping, or to avoid exceeding the yield stress of the object, at which point permanent deformation or failure occurs. Predicting deformation is also critical, especially when grasping containers. One may want to minimize the deformation on a box of crackers to avoid crushing the contents, or maximize the deformation on a bottle of ketchup to efficiently squeeze out the contents. 

Although knowledge of stress and deformation fields is useful, deriving closed-form solutions is intractable for general cases. Moreover, direct real-world measurement is extremely difficult  without cumbersome instrumentation.  Consequently, robotic grasping has historically leveraged rigid-body models, for which deformation is ignored and object state can be simply described by 6D pose and velocity~\cite{Murray1994,mason2001mechanics}. 

\begin{figure}
\centering
\includegraphics[width=\columnwidth]{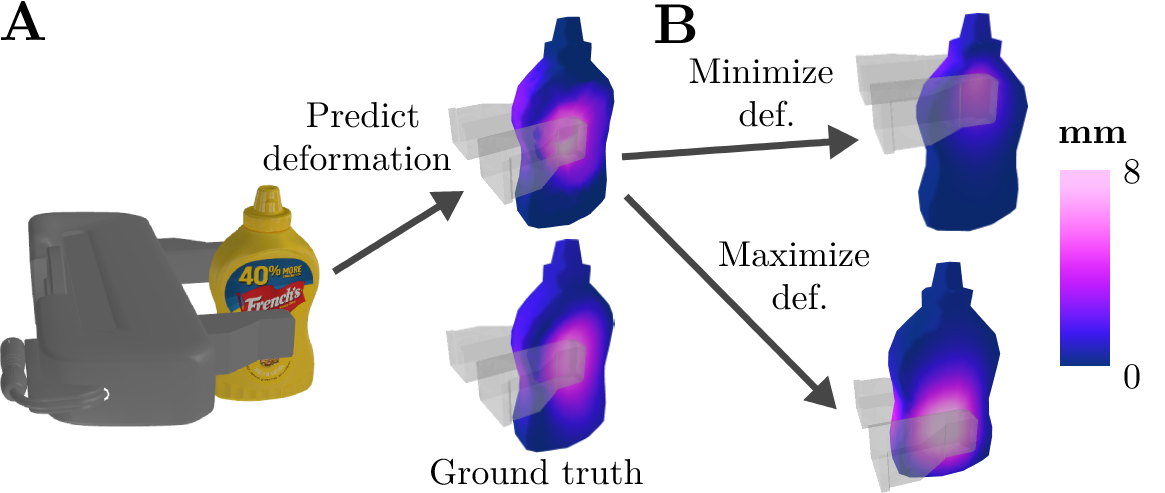}
\caption{\textbf{(A)} DefGraspNets predicts the stress and deformation fields from grasping an unseen object $1500$x faster than FEM, and \textbf{(B)} enables gradient-based grasp refinement to optimize these fields.}
\label{fig:front_fig}\vspace{-12pt}
\end{figure}

On the other hand, we can use deformable-object \textit{simulators} to access these quantities and plan grasps accordingly. However, such simulators rely on complex numerical models like the gold-standard 3D finite element method (FEM) \cite{YinScienceRobotics2021, Arriola2020FRAI}. Although FEM can simulate the result of any grasp on a deformable object \cite{Huang2022RAL}, each evaluation can take minutes on a CPU-based industry-standard simulator~\cite{Narang2021IJRR} and seconds on a GPU-based robotics simulator~\cite{Narang2021Latent}, which is prohibitively slow for online grasp planning. Moreover, while differentiable simulators enable gradient-based optimization for parameter estimation~\cite{Heiden2022AutoRob} and control optimization~\cite{Xu2022ICLR,Heiden2022AutoRob}, few studies have explored their application to robotic grasping, and they are typically slower than standard FEM.

We propose \emph{DefGraspNets}, a graph neural network that can enable grasp planning by predicting the stress and deformation fields resulting from grasps and allowing efficient optimization (Fig.~\ref{fig:front_fig}). We demonstrate that this network is 1) \textbf{fast}, with a ${\sim}1500$x speed-up compared to a GPU-based FEM simulator, 2) \textbf{accurate}, with stress and deformation fields consistent with ground-truth FEM, 3) \textbf{generalizable}, with reliable rankings of grasp candidates over unseen poses, elastic moduli, in-category objects, and out-of-category objects, and 4) \textbf{differentiable}, enabling gradient-based optimization for grasp refinement. Finally, we conduct pilot studies that verify agreement of DefGraspNets, trained purely in simulation, with real-world outcomes. Data and code can be found on our website\footnote{\url{https://sites.google.com/view/defgraspnets}}.


%% file: 2_related_work.tex
\section{Related Work}
Grasp planning has received significant attention in robotics~\cite{Grupen1991,Murray1994,sahbani2012overview,ciocarlie-2007-eigengrasps,Newbury2022arxis}. Recent works leverage learning-based approaches to enable fast planning and generalization to novel objects~\cite{lenz2015deep,mousavian20196,mahler2017dex,lu-ram2020-MultiFingeredGP,lundell-2021icra-fingan}. We focus on grasp planning for 3D deformable objects. Unlike rope or cloth, 3D deformables have dimensions of a similar magnitude along all 3 spatial axes and can undergo significant deformations along any of them~\cite{Huang2022RAL}.
We review grasp planning for 3D deformables, as well as methods for predicting stress and deformation fields via graph neural networks and differentiable simulation.

\subsection{Grasp planning for deformable objects}

Early works in grasp planning for deformable objects focused on finding \textit{stable} grasps of planar objects, under which the object's strain energy would be maximized without inducing plastic deformation~\cite{Goldberg2005IJRR, Jia2014IJRR}. This has since been extended to the 3D case, where novel time-dependent grasp quality metrics have been proposed to capture the evolution of contact states under deformation~\cite{Song2022RAL,Le2023IROS}.

Grasp planning for \textit{deformation} of thin-walled containers (e.g., boxes, bottles) has also been explored. Given a 3D geometric stiffness map of the object, a minimal deformation grasp can be planned by localizing contact at high-stiffness regions. This map can be generated in simulation, via real-world probing ~\cite{Xu2020ICRA}, or from 2D images of the object via generative adversarial networks~\cite{Makihara2022AR}. Grasp planning for \textit{stress} has also been demonstrated via simulation on quasi-rigid objects using the boundary element method~\cite{Pan2020ICRA}. Finally, grasp planning for additional metrics  can be performed with DefGraspSim, a 3D FEM-based grasp simulation framework \cite{Huang2022RAL}. For every grasp, it evaluates success, stability, stress, deformation, strain energy, and controllability.

These methods vary not only in the planning metric, but also in the type of computation required. Some require FEM simulation of the beginning of the interaction (e.g., just past initial contact)~\cite{Jia2014IJRR,Song2022RAL,Le2023IROS,Xu2020ICRA} or the full interaction~\cite{Pan2020ICRA,Huang2022RAL}, whereas others use neural networks~\cite{Makihara2022AR}. Yet, all of these planners can only evaluate or predict the outcome of a candidate grasp, and cannot optimize grasps through gradient-based methods.

\subsection{Graph neural networks for deformable-object interaction}
Graph neural networks (GNNs) have been used to efficiently learn  dynamics models for granular solids, deformable solids, and fluids  \cite{li2018learning, Sanchez020ICML, Ummenhofer2020ICLR, Pfaff2021ICLR, Shi2022RSS}. Inspiring our work, MeshGraphNets~\cite{Pfaff2021ICLR} used GNNs to learn accurate dynamics for deformable solids
using mesh-based representations, training from an industry-standard FEM solver. It predicted deformation and stress on a 3D deformable plate with kinematically-actuated colliding shapes and achieved evaluation speeds up to two orders of magnitude faster than the solver.
RoboCraft~\cite{Shi2022RSS} used GNNs to  learn how plasticine-like objects with particle representations deform under interaction with a robotic gripper, training from visual input. Whereas MeshGraphNets used forward passes through the networks to predict dynamics, RoboCraft also used backwards passes to perform gradient-based trajectory optimization, molding the plasticine into a desired shape.

Our work also utilizes a GNN as a surrogate simulator for dynamics predictions. Unlike MeshGraphNets, which uses $N$-step rollouts to predict a final state via intermediate steps, DefGraspNets performs direct, one-step predictions of the final state. One-step prediction ensures that gradients are only propagated once through the network rather than over tens or hundreds of steps, mitigating vanishing or exploding gradients \cite{lillicrap2019backpropagation}. Furthermore, we focus on quasistatic rather than dynamic grasping; the ability of multi-step rollouts to predict object and controller dynamics offers limited advantage. Our ablation study verifies that using single-step predictions in our setting performs better than multi-step predictions (c.f. Sec.~\ref{sec:ablation}).

Like RoboCraft, we design our network to include gripper actions in order to perform gradient-based optimization for grasp planning.
Unlike RoboCraft and MeshGraphNets, we use force rather than position commands for our actuators, as force commands are implemented in notable industrial grippers \cite{Lauzier2016Robotiq,Onrobot2019,Schunkgrippers} and are preferable for grasping (as opposed to applications like shape control). Gripper force determines whether the grasp will overcome the object's gravity, and gripper position cannot indicate force without additional knowledge (e.g., contact area, object stiffness). In addition, when grasping stiffer objects, position commands can induce high torques that can damage both the object and gripper. We also generalize our network to different elastic moduli, which was not explored in prior works. 

\subsection{Differentiable simulators}
Differentiable simulators for rigid and deformable bodies allow gradients of output variables (e.g., poses, velocities, or deformation fields of objects) to be computed with respect to input variables (e.g., control inputs or material parameters) \cite{warp2022, freeman2021brax, hu2019difftaichi, heiden2021neuralsim, werling2021fast, murthy2020gradsim, hu2019chainqueen}. Such simulators enable gradient-based optimization for control optimization~\cite{Xu2022ICLR,  murthy2020gradsim, huang2021plasticinelab, hu2019chainqueen, geilinger2020add}, parameter estimation \cite{Heiden2022AutoRob, geilinger2020add, hu2019chainqueen}, and inverse design \cite{xu2021end, hu2019chainqueen}.

There are 4 main strategies to realize a differentiable simulator or equivalent model: 1) finite-differencing a non-differentiable simulator, which has unfavorable $\mathcal{O}(n)$ scaling to an $n$-dimensional input space \cite{baydin2018automatic, margossian2019review}, 2) analytically or automatically differentiating a simulator that smoothly approximates spatial or kinetic discontinuities (e.g., penalty-based contact forces and smooth friction models \cite{Heiden2022AutoRob, geilinger2020add}, which may introduce inaccuracies or require tuning), 3) training a deep network with physically-based loss functions \cite{raissi2019physics, karniadakis2021physics}, which has seen limited use for contact dynamics \cite{pfrommer2020contactnets}, and 4) training a deep network on datasets from a non-differentiable simulator, primarily with graph-based inductive biases \cite{Battaglia2018arxiv, li2018learning, Sanchez020ICML, Pfaff2021ICLR}.

For our application, we aim to simulate robotic grasping of 3D deformable objects. Thus, we focus on gold-standard 3D FEM simulation of deformable objects with contact. For this application, strategy 2 has been explored in a handful of recent works \cite{Heiden2022AutoRob}, including differentiable projective dynamics \cite{Du2021TOG, qiao2021differentiable}. However, such simulators typically execute substantially slower than real-time (especially including a backward pass), and only one has realized differentiable FEM and contact modeled via the full nonlinear complementarity problem (NCP) \cite{horak2019similarities} with both static and dynamic friction \cite{qiao2021differentiable}.

In this work, we explore strategy 4, training for the first time on a GPU-accelerated robotics FEM simulator \cite{makoviychuk2021isaac} that addresses the full NCP \cite{macklin2019non} and has been experimentally validated across multiple studies~\cite{Huang2022RAL,Narang2021IJRR, Narang2021Latent}. To our knowledge, this effort also comprises the first application of such methods to robotic grasping of 3D deformable objects.

Strategy 2 has often been favored over strategy 4 due to the former's potential for generalizing to arbitrary physics \cite{Du2021TOG, qiao2021differentiable}. Nevertheless, we show for the first time that strategy 4, through judicious selection and scaling of training data, can indeed generalize to novel grasps, elastic moduli, in-category objects, and out-of-category objects. Furthermore, the trained networks   can execute 2 to 3 orders of magnitude faster than the reference simulator (i.e., faster than real-time).


%% file: 3_model.tex
\section{The DefGraspNets Model}
\begin{figure}
\centering
\includegraphics[width=\columnwidth]{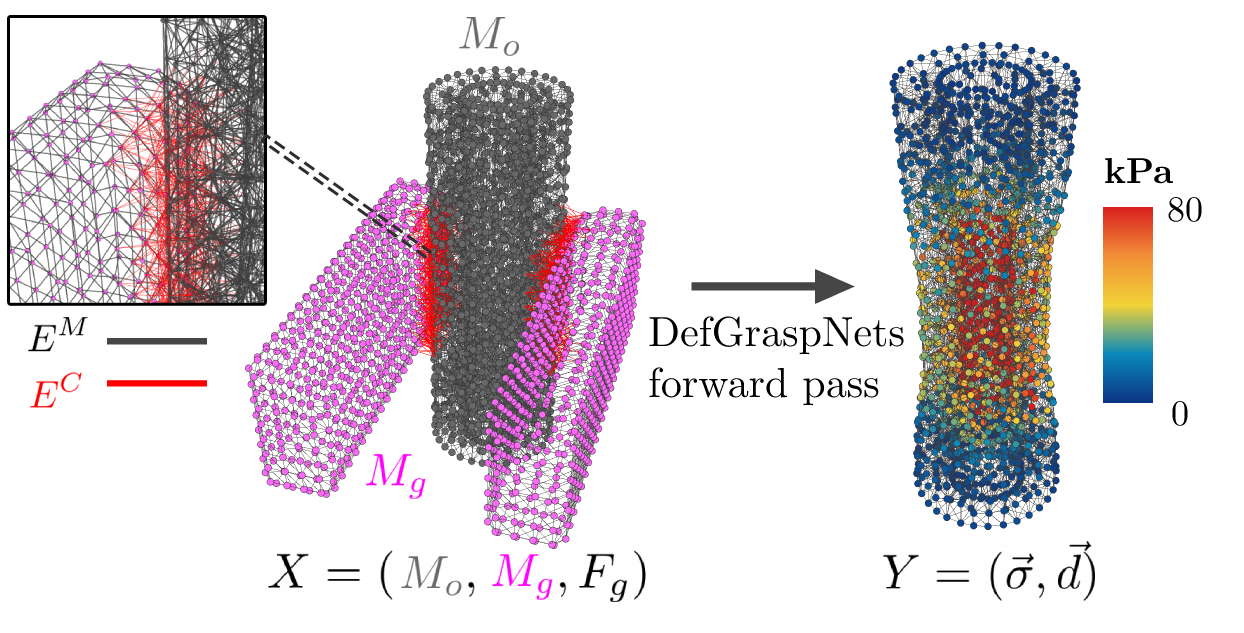}
\caption{Given a candidate grasp state $X$ consisting of an object mesh $M_o$, gripper mesh $M_g$, and grasp force $F_g$, DefGraspNets generates contact edges $E^C$ and predicts output $Y$ consisting of a stress field $\vec{\sigma}$ and deformation field $\vec{d}$ defined at each node of the object mesh.}
\label{fig:network_diagram}
\vspace{-18pt}
\end{figure}

Here, we explain the GNN structure of DefGraspNets, including the input and output representations.
We detail training data generation in Sec.~\ref{sec:training} and explain how we use DefGraspNets within a grasp planning algorithm in Sec.~\ref{sec:grasp-planning}.

\subsection{Summary of inputs and outputs}
DefGraspNets takes as input a \textit{candidate grasp state} $X = (M_o, M_g, F_g)$ comprising a mesh $M_o$ of a deformable object in its pre-contact state, a mesh $M_g$ of the gripper fingers upon initial contact\footnote{Although $M_g$ comprises two unconnected parts, we refer to it collectively as the ``gripper mesh."}, and a total normal grasp force scalar $F_g$. A mesh is a collection of vertices and undirected edges that connect them. For $M_o$, these vertices and edges form tetrahedral elements that define the volumetric geometry of the object. For $M_g$, the vertices and edges form triangular elements that define the surface geometries of the fingers.

DefGraspNets converts the candidate grasp state $X$ into a multigraph $G$ (Sec.~\ref{sec:multigraph}), mapping the gripper-object contact interactions onto a graph structure. The multigraph is fed into an \emph{Encode-Process-Decode} sequence~\cite{Pfaff2021ICLR,Sanchez020ICML,Battaglia2018arxiv}(Sec.~\ref{sec:encode}). DefGraspNets predicts the stress and deformation $Y = (\vec{\sigma}, \vec{d})$ at steady state, at all vertices of $M_o$ (Fig.~\ref{fig:network_diagram}). Please refer to Sec.~\ref{sec:training} for the formal definitions of these fields.

\subsection{Multigraph representation}\label{sec:multigraph}
The multigraph representation $G = (V, E^M, E^C)$ has nodes $V$ and undirected edge sets $E^M$ and $E^C$; each edge stores the indices of its two connected nodes. We list their features, and mark those that differ from \cite{Pfaff2021ICLR} with a $\star$ bullet.

\noindent\textbf{Nodes.} The nodes $V$ correspond to the vertices of $M_o$ and $M_g$. Each node $v_i$ has a feature vector consisting of
\begin{itemize}[leftmargin=2ex]
    \item A 3-element one-hot vector for node type (i.e., part of $M_g$, $M_o$ surface, or $M_o$ interior)
    \item The 3D Cartesian position of the node
    \item[{\large$\star$}] A 3D unit vector in the gripper closing direction. This is nonzero only for gripper nodes and informs the network which direction the grippers are closing. 
\end{itemize}

\noindent\textbf{Mesh edges.}
The mesh edges $E^M$ correspond to the edges of $M_o$ and $M_g$. Each mesh edge $e^M_{ij}$ connects nodes $v_i$ and $v_j$ of the same type. Its feature vector consists of
\begin{itemize}[leftmargin=2ex]
    \item The 3D Cartesian displacement vector from $v_i$ to $v_j$
    \item The scalar Euclidean distance between $v_i$ and $v_j$
    \item[{\large$\star$}] The scalar elastic modulus $E$ of the deformable object. This is nonzero only for edges belonging to the object. 

\end{itemize}

\noindent\textbf{Contact edges.}
The contact edges $E^C$ connect object and gripper nodes and are computed based on proximity at initial contact. Each edge $e^C_{ij}$ is formed between a pair of nodes $v_i$ and $v_j$ that have different node types and are closer than hyperparameter $\epsilon$. The edge's feature vector comprises
\begin{itemize}[leftmargin=2ex]
    \item The 3D Cartesian displacement vector from $v_i$ to $v_j$
    \item The scalar Euclidean distance between $v_i$ and $v_j$
    \item[{\large$\star$}] The normalized grasp force $F_g^C$, which is the total grasp force $F_g$ divided by the number of contact edges $|E^C|$. 
    
\end{itemize}

 \subsection{Encoder, processor, \& decoder architectures}\label{sec:encode}
First, all feature vectors associated with the nodes $V$, mesh edges $E^M$, and contact edges $E^C$ are encoded into a common latent space with 3 respective multilayer perceptrons (MLPs). Then, $L$ message-passing blocks with 3 separate MLPs per block sequentially aggregate and process information from adjacent nodes and edges. Finally, a decoder MLP takes the processed nodal features in the latent space and jointly outputs the predicted stress and Cartesian displacement per node in real units ($\units{Pa}$ and $\units{m}$). Full details of the Encode-Process-Decode sequence can be found in \cite{Pfaff2021ICLR}.


%% file: 4_training.tex
\section{Data Generation and Model Training}\label{sec:training}
We now describe our simulation-based approach to training DefGraspNets.
We design a set of 60 object primitive models as a high-level abstraction of real-world geometries grouped into geometric categories (e.g., cuboids, cylinders, ellipsoids, annuli), and instances within each category have different dimensions and aspect ratios. Our dataset also includes a set of 11 of fruits and vegetables (e.g., apples, eggplants, potatoes) based on 3D scans \cite{ybjDataset}. Tetrahedral volume meshes are generated for each deformable object using fTetWild~\cite{ftw}. Triangular surface meshes are generated for the gripper fingers using Onshape.

For each pre-contacted object mesh $M_o$, 100 grasps are generated using an antipodal sampler~\cite{EppnerISRR2019} wherein randomly-sampled surface points define gripper contact points, surface normals define grasp axes, and 4 rotations are regularly drawn about each grasp axis. These 100 grasps correspond to 100 gripper meshes $M_g$. Each grasp is evaluated using the DefGraspSim\cite{Huang2022RAL} simulation framework (built upon Isaac Gym\cite{makoviychuk2021isaac} and the FleX FEM solver\cite{macklin2019non}) with the Franka parallel-jaw gripper. DefGraspSim evaluates the stress and deformation fields of the deformable object during grasping. 

Given an object-grasp pair $(M_o, M_g)$ in DefGraspSim, the gripper applies a linearly increasing amount of force on the object until $F_g^{max} = 15$\units{N} is reached in a zero-gravity environment.\footnote{For the elastic moduli examined ($1e^4 \leq E \leq 1e^7$~\units{Pa}), $15$\units{N} was observed to induce substantial stress and deformation; gravity was ignored due to having negligible effect on stress and deformation compared to contact forces.} This  force was achieved by directly commanding DOF torque applied at the gripper joints. The values of the stress ($\vec{\sigma}$) and deformation fields ($\vec{d}$) at all object vertices are saved over 50 evenly-spaced substeps throughout the entire grasping trajectory. 
Formally, our dataset $D$ is composed of input-output pairs, each consisting of a candidate grasp pose $X_i$ and corresponding set of fields $Y_i$, that is, $D=\{X_i=(M_g, M_o, F_g), Y_i=(\vec{\sigma},\vec{d})\}_{i=1}^N$, where $0 \leq F_g \leq 15$. Dataset $D$ has $N = \#~objects \times 100 \times 50=3.55e5$ unique points. Because our network performs one-step predictions of the final state and is ideal for quasistatic interactions, all unstable grasps involving chaotic dynamics are not included in $D$.

The values of the stress field $\vec{\sigma}$ at all object vertices are computed as follows: first, the second-order stress tensor at each tetrahedral element of $M_o$ is acquired from DefGraspSim. The stress tensors at each vertex are calculated by averaging the stress tensors at all adjacent elements. Each stress tensor is then converted to the scalar von Mises stress (i.e., the second invariant of the deviatoric stress), which is widely used to quantify whether a material has yielded \cite{timoshenko2010}. The values of the deformation field $\vec{d}$ are defined simply as the distance between the positions of the pre-contacted vertices of $M_o$ and their positions under gripper force $F_g$. 

Contact edges $E^C$ are formed based on the threshold $\epsilon = 5$mm. Our networks are trained with a decaying learning rate from $5e^{-5}$ to $1e^{-6}$ over 25 epochs and a batch size of $1$. A latent size of $128$ and $L=15$ message passing steps are used, where all MLPs have 2 hidden layers. Loss is defined as the sum of the MSE of stress and deformation over all nodes. On a single RTX 3090 GPU, the network trains at approximately 1600 steps per minute.  

%% file: 5_grasp_planning.tex
\section{Grasp Planning}\label{sec:grasp-planning}
We demonstrate DefGraspNets as a grasp planner, where both gradient-free (i.e., evaluation of sampled grasps) and gradient-based refinement methods can be used to find an optimal grasp. We define $Q$ as the optimization objective, which is any backwards pass-differentiable measure of the predicted deformation and/or stress fields (e.g., mean deformation, smooth differentiable approximation of maximum stress implemented in modern deep learning libraries).

\subsection{Evaluation of sampled grasps}
First, DefGraspNets supports online sampling-based grasp planning. For an unseen object, forward passes of DefGraspNets can be used to evaluate $Q$ for 100 random antipodal grasps with parallel batches of size 5 in 7.3 seconds. In comparison, DefGraspSim requires approximately 3 hours to evaluate 100 grasps, which is ${\sim}1500$x slower. 

The best grasp pose is identified as $T^* = \arg\min_{T \in T_s} Q(T; M_o)$, where $T$ is a 6D rigid transformation applied to a constant initial state of the gripper $M_g^{0}$ wherein both fingers are maximally open. Any valid $M_g$ can be fully defined by $T$ and joint states $\vec{p}_g \in \mathbb{R}^2$ that determine how much each finger closes in order to contact $M_o$. These joint states $\vec{p}_g$ are calculated analytically by projecting the vertices of $M_o$ onto the gripper faces, backprojecting the vertices within each face, and computing the minimum perpendicular distance over these vertices (i.e., the minimum contact distance) per finger.

\subsection{Grasp refinement}
Unlike existing deformable object planners, DefGraspNets' differentiability enables gradient-based refinement of a grasp pose to optimize $Q$. Starting from an initial grasp pose $T_{init}$, we perform gradient updates in the direction of $\sfrac{\partial{Q}}{\partial T}$ to achieve a refined $T$ using backtracking line search \cite{NoceWrig06} and simulated annealing \cite{Laarhoven87}. With 12 refinement steps per grasp, refining 100 initial grasps requires approximately $8$ minutes. A comparable time does not exist for DefGraspSim, as it is not differentiable.

%% file: 6_prediction_results.tex
\section{Prediction Results}
We test DefGraspNets' predictions of the ranking of grasps with respect to their mean stress and deformation values by quantifying the respective Kendall's $\tau$ rank correlation coefficients ($\tau_{s}$ and $\tau_{d}$).\footnote{Kendall's $\tau$ was chosen over Spearman's $\rho$ for its comparative robustness (i.e., smaller gross error sensitivity).} We answer the following questions for 4 levels of generalization:

\begin{figure*}[ht]
     \centering
     \begin{subfigure}[b]{.3\textwidth}
         \centering
         \includegraphics[scale=0.23, trim={0cm 0cm 0cm 0cm},clip]{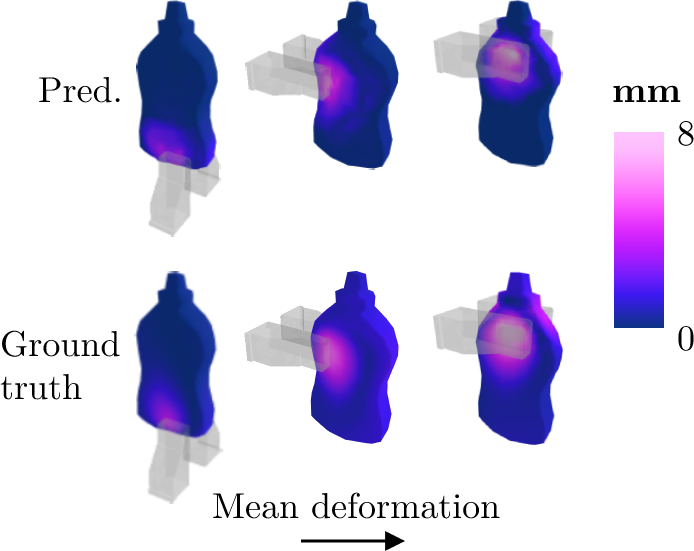}
         \caption{Mustard bottle, $F_g = 12
         \units{N}$, $E = 1e7$}
         \label{fig:pred_mustard}
     \end{subfigure}
     \hfill
     \begin{subfigure}[b]{.3\textwidth}
         \centering
         \includegraphics[scale=0.23, trim={0cm 0cm 0cm 0cm},clip]{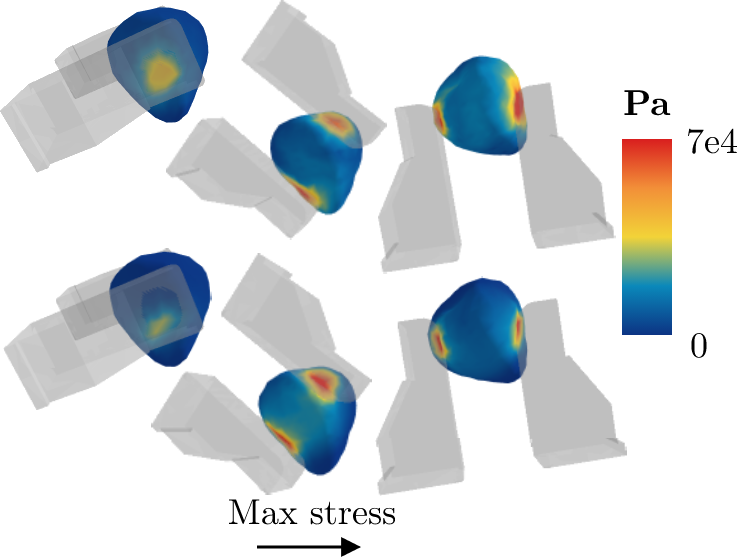}
         \caption{Strawberry, $F_g = 6\units{N}$, $E = 5e4$}
         \label{fig:pred_strawberry}
     \end{subfigure}
     \hfill
     \begin{subfigure}[b]{.3\textwidth}
         \centering
         \includegraphics[scale=0.23, trim={0cm 0cm 0cm 0cm},clip]{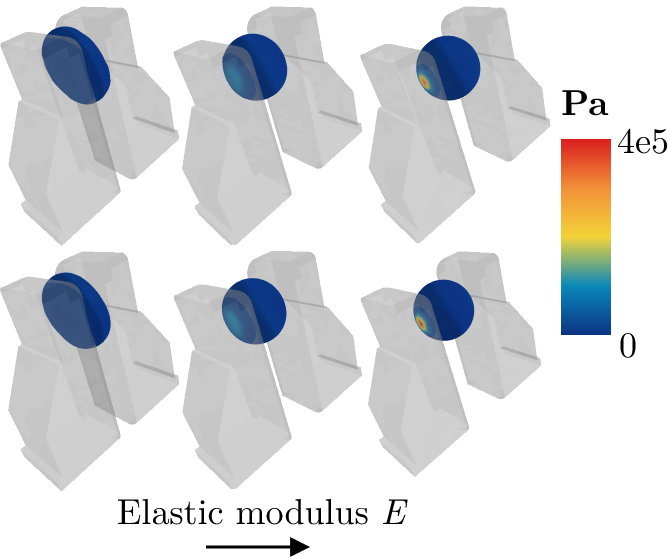}
         \caption{Sphere, $F_g = 5\units{N}$, $E \in [5e5, 1e6, 5e6]$}
         \label{fig:pred_lemon}
     \end{subfigure}
     \hfill
        \caption{A) Predicted and ground-truth deformation fields for a mustard bottle subject to grasps inducing increasing mean deformation, B) Predicted and ground-truth stress fields for a strawberry subject to grasps inducing increasing maximum stress, and C) Predicted and ground-truth stress fields for a sphere of increasing elastic moduli subject to identical grasps (deformation can be seen in resulting shape).
        }
        \label{fig:predictions}
\end{figure*}

\begin{enumerate}[leftmargin=2ex]
    \item Can DefGraspNets rank unseen grasps when trained on other grasps on the same object? (Ans: Yes. For an 80-20 train-test split over grasps on the same object, we get an average $\tau_{s} = 0.78$ and $\tau_d = 0.66$ over $15000$ unseen $X_i$.)
    \item Can DefGraspNets generalize to unseen elastic moduli $E$ on the same object? (Ans: Yes. For a 7-3 train-test split over unique $E$ for grasps on the same object, we get an average $\tau_{s} = 0.81$ and $\tau_d = 0.72$ over $15000$ unseen $X_i$.)
    \item Can DefGraspNets generalize to unseen primitive objects within the same geometric category? (Ans: Yes. For a 5-1 train-test split over unique objects, we get an average $\tau_{s} = 0.48$ and $\tau_d = 0.54$ over $15000$ unseen $X_i$.)
    \item Can DefGraspNets generalize to unseen real-world objects? (Ans: Yes. Moreover, we generate useful predictions even when training on a small number of objects, as long as the train geometries are relevant to the test geometry as quantified by a low Chamfer distance. See Table~\ref{tab:g4}, which also reports the mean absolute error (\textit{MAE})). 
\end{enumerate}



Full visualizations of predicted field quantities for the 4th (i.e., most challenging) generalization level is shown in Fig.~\ref{fig:predictions} on an unseen mustard bottle and unseen strawberry, as well as for the 2nd generalization level on a sphere.

\begin{table*}[!htp]\centering
\caption{Generalization to unseen real-world objects. Gray cells denote the best values per column. Train sets each contain only 5 objects; the ``All" group contains all 15. The $d_C$ column measures the best Chamfer distance between the test geometry and the train geometries. Lower $d_C$ implies geometric similarity between the train and test objects, and corresponds to more favorable MAE and $\tau$ during prediction.}\label{tab:g4}
\scriptsize
\begin{tabular}{lrrrrrrrrrrrrrrrr}\toprule
&\multicolumn{5}{c}{\textbf{Mustard bottle}} &\multicolumn{5}{c}{\textbf{Lemon half}} &\multicolumn{5}{c}{\textbf{Strawberry}} \\\cmidrule{2-16}
\multirow{2}{*}{\textbf{Train set}} &\multirow{2}{*}{$d_C$ [mm] ↓} &\multicolumn{2}{c}{Deformation [mm]} &\multicolumn{2}{c}{Stress [kPa]} &\multirow{2}{*}{$d_C$  ↓} &\multicolumn{2}{c}{Deformation} &\multicolumn{2}{c}{Stress} &\multirow{2}{*}{$d_C$  ↓} &\multicolumn{2}{c}{Deformation} &\multicolumn{2}{c}{Stress} \\\cmidrule{3-6}\cmidrule{8-11}\cmidrule{13-16}
& &MAE ↓ &$\tau_d$ ↑ &MAE ↓ &$\tau_s$ ↑ & &MAE ↓ &$\tau_d$ ↑ &MAE ↓ &$\tau_s$ ↑ & &MAE ↓ &$\tau_d$ ↑ &MAE ↓ &$\tau_s$ ↑ \\\midrule
Group 1 &\cellcolor[HTML]{efefef}5.57 &\cellcolor[HTML]{efefef}0.71 &\cellcolor[HTML]{efefef}0.62 &\cellcolor[HTML]{efefef}2.92 &\cellcolor[HTML]{efefef}0.56 &3.57 &4.82 &0.20 &2.15 &0.31 &3.27 &0.42 &0.09 &6.41 &0.58 \\
Group 2 &6.77 &0.74 &0.20 &4.72 &0.45 &\cellcolor[HTML]{efefef}3.30 &3.98 &\cellcolor[HTML]{efefef}0.50 &\cellcolor[HTML]{efefef}1.30 &\cellcolor[HTML]{efefef}0.43 &2.61 &0.30 &0.29 &2.85 &0.54 \\
Group 3 &6.07 &0.73 &-0.31 &4.57 &0.45 &4.50 &4.28 &-0.03 &2.63 &0.19 &\cellcolor[HTML]{efefef}2.46 &0.31 &-0.05 &2.79 &0.64 \\
All &\cellcolor[HTML]{efefef}5.57 &0.73 &0.60 &3.66 &\cellcolor[HTML]{efefef}0.56 &\cellcolor[HTML]{efefef}3.30 &\cellcolor[HTML]{efefef}3.98 &0.43 &1.36 &\cellcolor[HTML]{efefef}0.43 &\cellcolor[HTML]{efefef}2.46 &\cellcolor[HTML]{efefef}0.30 &\cellcolor[HTML]{efefef}0.39 &\cellcolor[HTML]{efefef}2.68 &\cellcolor[HTML]{efefef}0.66 \\
\bottomrule
\end{tabular}
\vspace{-5pt}
\end{table*}

%% file: 7_grasp_planning_results.tex
\section{Grasp Planning Results}
We demonstrate DefGraspNets as a grasp planner on $3$ unseen objects (a mustard bottle, a lemon, and a strawberry) from existing datasets~\cite{Calli2015ICAR,TurboSquid} with real-world elastic moduli. First, we perform evaluation of sampled grasps. On each unseen object, $100$ random grasps $T_r$ are generated, and the optimization metric $Q(T)$ is evaluated for each $T \in T_r$ via the forward pass of DefGraspNets. Of the $100$ grasps, we select the $10$ grasps that are predicted to yield the lowest $Q$ (``threshold low" grasps), as well as $10$ grasps that are predicted to yield the highest $Q$ (``threshold high" grasps). We also randomly select $10$ other grasps from the remaining $80$ grasp candidates as a baseline. These $30$ grasps are then evaluated within the ground-truth simulator DefGraspSim. 

DefGraspNets is a reliable predictor of minimal- and maximal-$Q$ grasps on the unseen objects, with $88\%$ of these threshold-low and high grasps belonging to the set of 30 lowest and highest ground-truth-$Q$ grasps, respectively. 

\begin{figure}
\centering
\includegraphics[width=\columnwidth]{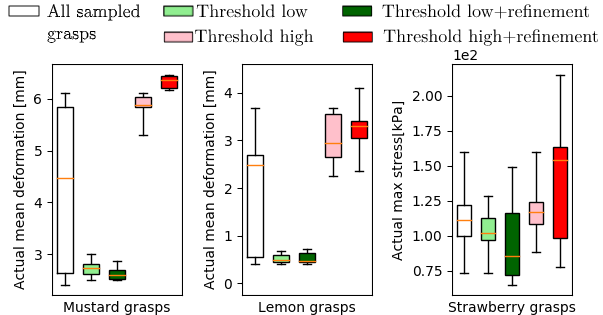}
\caption{Box plots for $5$ groups of grasps for each unseen object: 1) all grasps, 2) threshold low grasps from sampling only, 3) threshold low grasps after refinement, 4) threshold high grasps from sampling only, and 5) threshold high grasps after refinement. The $y$-axis is the ground-truth $Q$ value of these grasps as computed in DefGraspSim.}
\label{fig:boxplots}
\end{figure}

Subsequently, we perform gradient-based grasp refinement on the threshold-low and threshold-high grasps to further reduce and increase $Q$, respectively. For each object, box plots in Fig.~\ref{fig:boxplots} visualize the distribution of ground-truth $Q$ values for $5$ groups of grasps: all sampled grasps, threshold-low grasps, threshold-low grasps after refinement, threshold-high grasps, and threshold-high grasps after refinement. In all cases, not only do threshold-high and low grasps from DefGraspNets yield substantially different ground-truth $Q$ values, but refinement increases their polarity as desired. 

\begin{figure}
\centering
\includegraphics[width=0.83\columnwidth]{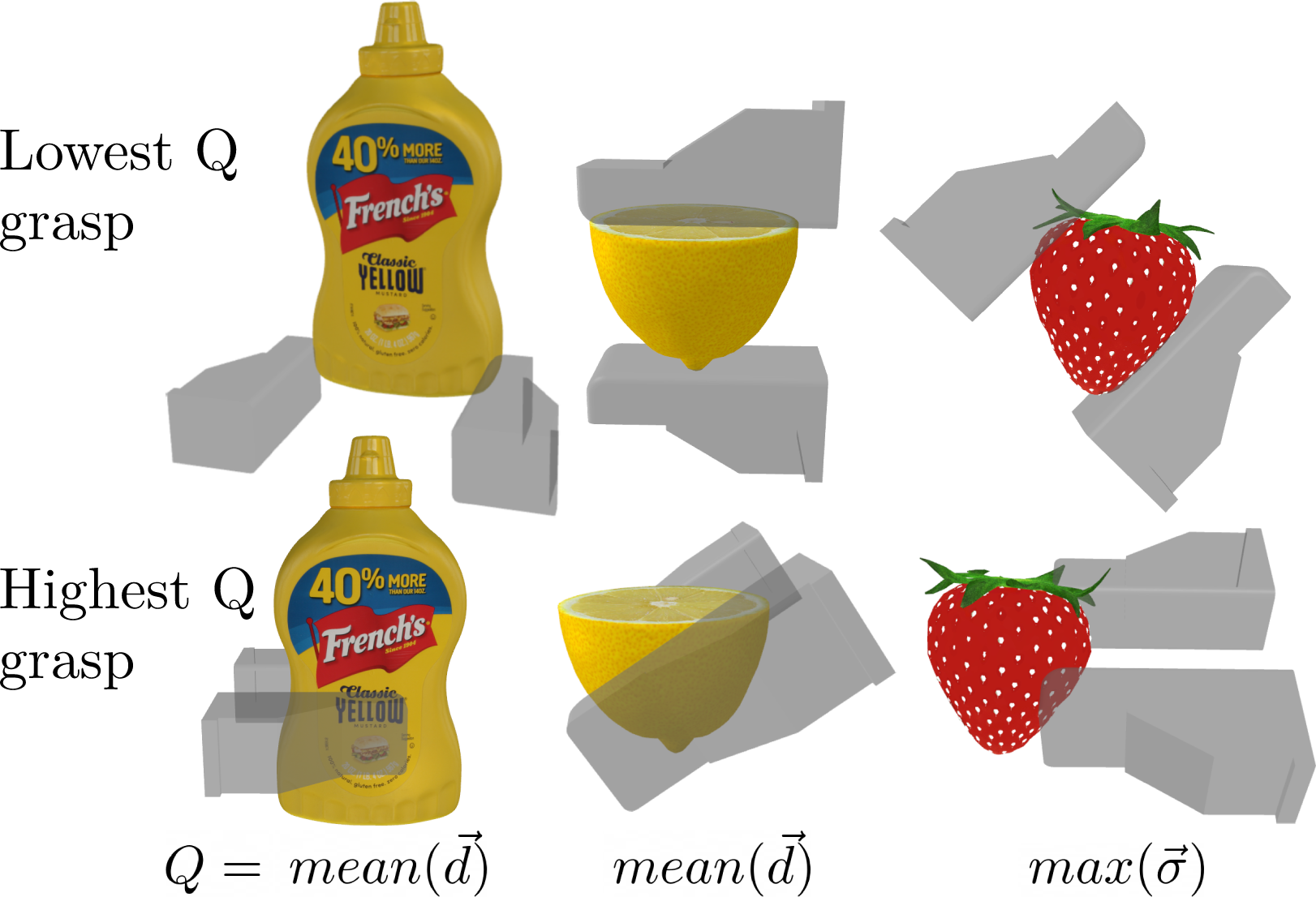}
\caption{Highest- and lowest-$Q$ grasps for the mustard bottle, lemon, and strawberry generated by the sample-and-refine procedure.}
\label{fig:minmax}
\vspace{-18pt}
\end{figure}

\begin{figure}[h]
\centering
\includegraphics[width=\columnwidth]{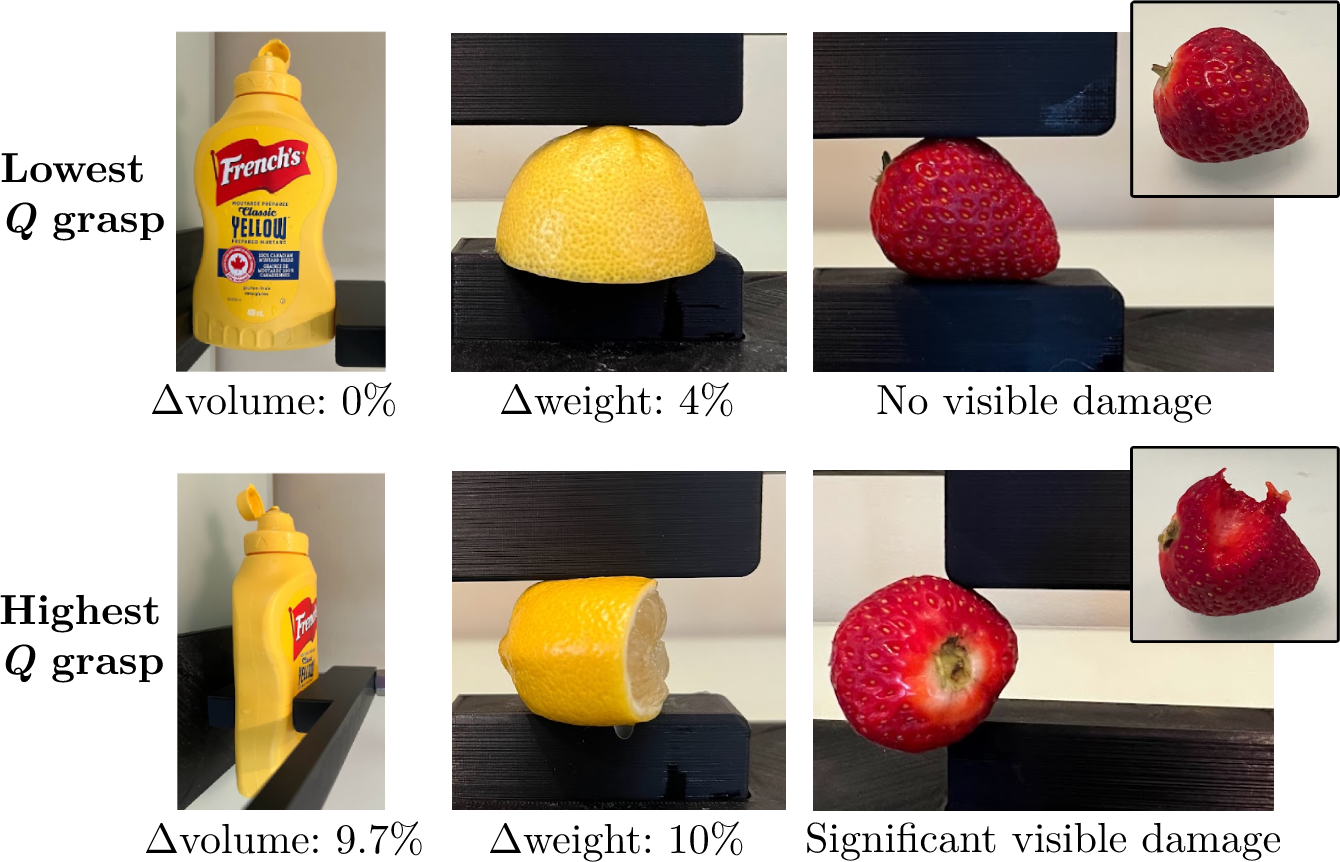}
\caption{Validation of grasps from Fig.~\ref{fig:minmax} using a Franka-based gripper gravitationally loaded under $15\units{N}$. For the bottle and lemon, deformation is measured by proxy (change in volume and weight). For the strawberry, only the highest-$Q$ grasp imparts damage.}
\label{fig:real_world}
\vspace{-14pt}
\end{figure}

The highest- and lowest-$Q$ grasps generated by the sample-and-refine grasp planning procedure are shown in Fig.~\ref{fig:minmax}. These grasps align with physical reasoning (e.g., the highest-deformation grasps on the bottle and lemon compress the directions of lowest geometric stiffness; the highest-stress grasp on a strawberry concentrates force on minimal area). These grasps are also validated in the real world in Fig.~\ref{fig:real_world}.

%% file: 8_ablation.tex
\section{Ablation Studies}\label{sec:ablation}
We run several ablation studies on our network architecture design. Table~\ref{tab:ablation} lists key design variables in DefGraspNets, with our selected conditions in bold. We compare our baseline model with 5 other trained models, each of which differ from baseline by exactly one condition.  We compare performance on a fixed test set and report the Kendall's $\tau$ metric for mean $\vec{d}$ and $\vec{\sigma}$. 
We address the following questions:
\begin{itemize}[leftmargin=2ex]
    \item Does jointly predicting stress and deformation outperform using two separate networks to predict these quantities? (Ans: The $\tau$ metric is comparable in both cases, likely because stress and deformation are coupled through the equations of elasticity. Thus, training two networks would be strictly disadvantageous computationally, c.f. V1.)
    \item Does one-step prediction outperform multi-step prediction? (Ans: Yes, when predicting deformation. Otherwise, both are comparable when predicting stress, c.f. V2. In MeshGraphNets, multi-step prediction does not accumulate significant deformation errors  because the trajectory of the actuators is exactly controlled. In DefGraspNets, gripper force is commanded; the positions of \textit{both} $M_o$ and $M_g$ are predicted and subject to accumulating errors.) 
    \item Should $F_g$ be normalized by the number of contact edges? (Ans: Yes. This aligns with simulation, in which the total force is the sum of forces at all contact points, c.f. V3.)
    \item Should force features $F_g^C$ be assigned to contact edges or to gripper nodes? (Ans: The network is able to incorporate this information equally well, c.f. V4.)
\end{itemize}

\begin{table}[!htp]\centering
\caption{Ablation study variables and conditions. Our DefGraspNets network conditions are in bold. Best conditions are in gray.}\label{tab:ablation}
\scriptsize
\begin{tabular}{lrrrr}\toprule
\textbf{Variable} &\textbf{Condition} &\textbf{$\tau_d$↑} &\textbf{$\tau_s$↑} \\\midrule
\multirow{3}{*}{V1. Num. outputs} &\textbf{Def. and stress} &\cellcolor[HTML]{efefef}0.61 &0.82 \\
&Def. only &0.57 & \\
&Stress only & &\cellcolor[HTML]{efefef}0.84 \\ \midrule
\multirow{2}{*}{V2. Prediction type} &\textbf{One-step predictions} &\cellcolor[HTML]{efefef}0.61 &\cellcolor[HTML]{efefef}0.82 \\
&Multi-step &0.37 &0.70 \\ \midrule
\multirow{2}{*}{V3. Value of $F_g^W$} &\textbf{Distributed, $\mathbf{\sfrac{F_g}{|E^W|}}$} &\cellcolor[HTML]{efefef}0.61 &\cellcolor[HTML]{efefef}0.82 \\
&Non-distributed $F_g$ &0.33 &0.51 \\ \midrule
\multirow{2}{*}{V4. Assignment of $F_g$} &\textbf{On world edges $E^W$} &\cellcolor[HTML]{efefef}0.61 &\cellcolor[HTML]{efefef}0.82 \\
&On all nodes $V$ &0.58 &0.74 \\
\bottomrule
\end{tabular}
\vspace{-14pt}
\end{table}

%% file: 10_discussion.tex
\section{Discussion and Future Work}
We present DefGraspNets, a differentiable GNN-based model for FEM simulation of 3D stress and deformation fields. We demonstrate that training DefGraspNets on a diverse set of grasps on primitive geometries enables effective prediction and grasp planning on unseen, real-world geometries. DefGraspNets enables not only fast evaluation of sampled candidate grasps ($1500$x faster than GPU-accelerated FEM), but also gradient-based refinement of these grasps to optimize field quantities (e.g., max stress and mean deformation). We verify the effectiveness of optimized grasps on novel objects both in the ground-truth FEM simulator and in the real world. 

To expand DefGraspNets for use in downstream manipulation tasks such as food preparation or robotic surgery, prediction of additional quantities should be explored. These may include stability during transport and deformation and flow under reorientation and gravity. 
Furthermore, as FEM simulators evolve, DefGraspNets can be retrained to predict  soft-soft contact or heterogeneous material responses. 

Developing data augmentation techniques for \textit{meshes} may enable vast dataset scaling from a minimal set of object models, further strengthening our ability to generalize to unseen objects. 
In addition, as our network is differentiable, techniques such as Stein variational gradient descent \cite{NIPS2016_b3ba8f1b} and stochastic gradient Langevin dynamics \cite{Welling2011Lan} may allow us to provide probabilistic, multi-modal \textit{distributions} of optimal grasps. Finally, architecture optimization (e.g., sparsity acceleration \cite{Choy2019CVPR}) may lead to even faster performance. 

DefGraspNets contributes the first differentiable approach to deformable grasp planning capable of predicting and optimizing stress and deformation fields on novel objects. We believe this coupling of fast prediction of field quantities with a differentiable model will enable a wide range of users to apply deformable grasp planning to their target domains.

\section{Acknowledgment}
\noindent We thank Miles Macklin and Eric Heiden for simulation expertise; Ankur Handa for network design advice; and Balakumar Sundaralingam and Clemens Eppner for insightful discussions.